\documentclass{article}

\usepackage{microtype}
\usepackage{graphicx}
\usepackage{subfigure}
\usepackage{enumerate}
\usepackage{scrextend}
\usepackage{booktabs}

\usepackage{hyperref}

\usepackage[accepted]{Conferencestyle}

\icmltitlerunning{Computational principles of intelligence}

\begin{document}

\twocolumn[
\icmltitle{Computational principles of intelligence: \\
            learning and reasoning with neural networks}

\icmlsetsymbol{equal}{*}

\begin{icmlauthorlist}
	\icmlauthor{Abel Torres Montoya}{to}
\end{icmlauthorlist}

\icmlaffiliation{to}{Dataveras}

\icmlcorrespondingauthor{}{abel@dataveras.com}

% You may provide any keywords that you
% find helpful for describing your paper; these are used to populate
% the "keywords" metadata in the PDF but will not be shown in the document
\icmlkeywords{Machine Learning, Artificial Intelligence, Reasoning}

\vskip 0.3in
]

% this must go after the closing bracket ] following \twocolumn[ ...

\printAffiliationsAndNotice{}  % leave blank if no need to mention equal contribution

\begin{abstract}
Despite significant achievements and current interest in machine learning and artificial intelligence, the quest for a theory of intelligence, allowing general and efficient problem solving, has done little progress. This work tries to contribute in this direction by proposing a novel framework of intelligence based on three principles. First, the generative and mirroring nature of learned representations of inputs. Second, a grounded, intrinsically motivated and iterative process for learning, problem solving and imagination. Third, an ad hoc tuning of the reasoning mechanism over causal compositional representations using inhibition rules. Together, those principles create a systems approach offering interpretability, continuous learning, common sense and more. This framework is being developed from the following perspectives: as a general problem solving method, as a human oriented tool and finally, as model of information processing in the brain.
\end{abstract}

\section{Introduction}
\label{submission}
Machine learning (ML) and artificial intelligence (AI) have become a driving force for new data processing applications and are positioned to radically transform many aspects of society, technology and science given that more general methods of problem solving are found.

Almost a decade after the first applications leading to the current unprecedented interest in ML and AI, there is also a growing debate about the limitations of existing solutions for building artificial general intelligence (AGI) \cite{Jordan18,Marcus18,Pearl18}. Key missing functionalities for AGI include: common sense reasoning, unsupervised learning and zero-shot learning among others. 

Several research groups are addressing the missing capabilities and the generalization problem by doing gradual improvements over the existing solutions. This way, we see effort directed to interpretability \cite{Samek17,Zhang18,Frosst17}, visual reasoning \cite{Perez17}, zero-shot learning \cite{Lake15,Santoro16}, intuitive physics \cite{Battaglia16,Fischer16,Chang16,Fragkiadaki15,Kansky17}, catastrophic forgetting \cite{Kirkpatrick17} and other issues. 

Deep learning (DL) continues to be the dominant paradigm, combined with extensive parameter tuning \cite{Kukacka17} and other task-dependent techniques: attention \cite{ Yoo16}, hierarchical models \cite{Ha18}, reinforcement learning  \cite{Nachum18, Peng18}, adversarial networks \cite{Zhang18a},  autoencoders \cite{Oord17}, meta-learning  learning \cite{Finn17} and more.

Despite the progress in the mentioned topics, there has been scarce progress towards general intelligence solutions. Generalization properties of DL are still poorly understood \cite{Zhang16,Recht18}, brittle \cite{Azulay18,Rosenfeld18,Alcorn18} and vulnerable to attacks \cite{Brown17,Li18} even in simple pattern detection scenarios. This leads to questioning whether the input-output paradigm based on loss function minimization and differentiable representations is sufficient for implementing reasoning and general problem solving.

In alignment with Lake et. al. \yrcite{Lake16} and Gershman et. al. \yrcite{Gershman15} we believe that the research on core mechanisms and theory of intelligence will be crucial for making progress in general intelligence implementations. Looking for those core principles, here we take a fresh look at intelligence, from the definition and verification till the high level reasoning processes. 

A systems approach is taken in relation to the functioning of intelligence: different mechanisms (perception, learning, reasoning or creativity) have a substantial interdependence and have to be developed simultaneously in the same framework. Ideally, we want to find a reduced number of fundamental principles from which multiple aspects of intelligence could be generated.

The proposed framework uses intrinsic motivation, mirrored representations and action inhibition to build those few core principles. The resulting mechanism contrasts with current deep learning and reinforcement learning (RL) paradigms in several ways:
\begin{itemize}
\item The initial state of the network is empty (versus multilayer connections in DL).
\item Information is processed incrementally without a dedicated training phase.
\item Learning and problem solving are equivalent for the system and handled with the same mechanisms.
\item Information processing takes place in loops and not in a feedforward way.
\item No manual configuration (i.e. loss function or rewards) is used to guide the system towards the goal.
\item The information processing capabilities of the network are adjusted to the context during computations, while in DL they are fixed after training.
\end{itemize}

We start by revisiting some of the AI terminology and the evaluation criteria for intelligence, defining the key aspects to be checked in the system. The framework is contrasted versus the expected properties as a problem solver, as a tool and as a model of information processing in the brain. Key properties of the system are interpretability thanks to abstract representations, unsupervised learning based on self-motivation and reasoning based on causal interactions.

The fact that learning and problem solving are unified under a common mechanism simplifies the complexity of the solution. Such simplicity plus the obtained generalization properties and biological plausibility, favors the argument of the proposal as general principles. In this model inhibition and causality enable logic operations and reasoning in the context of neural networks; which are missing aspect in DL solutions.

We believe that only by working simultaneously in the theory and the implementation of intelligent systems it will be possible to make relevant progress towards generalization. We expect to see more proposals in this direction increasing the synergy between AI models, practical implementations and neuroscience. 
\section{Reconsidering the Definition and Evaluation of Intelligence}
A broadly accepted definition of intelligence is the \emph{“ability to achieve goals in a wide range of environments”} \cite{Legg07}. This behavioral evaluation seems intuitively reasonable but has some ambiguity when used as a practical tool. Interpreted literally, it may lead to the creation of specialized behavior emulators aimed at fooling the tests.

The same situation happens if we just count the number of successful environments without considering the underlying data processing mechanisms. For example, a system that can solve problems in two areas (e.g. image recognition and text-to-speech translation) can be considered more intelligent than another that can only do one, given the same performance. In practice, the system performing more tasks may just have a large amount of hardcoded input-output associations while the other is analyzing potential solutions to the problem. To deal with this, here we use the following definition.

\textbf{Definition} \emph{Intelligence is the capacity of the system to create new knowledge/skills in order to solve a task}.
 \begin{figure}[ht]
	\vskip 0.2in
	\begin{center}
		\centerline{\includegraphics[width=\columnwidth]{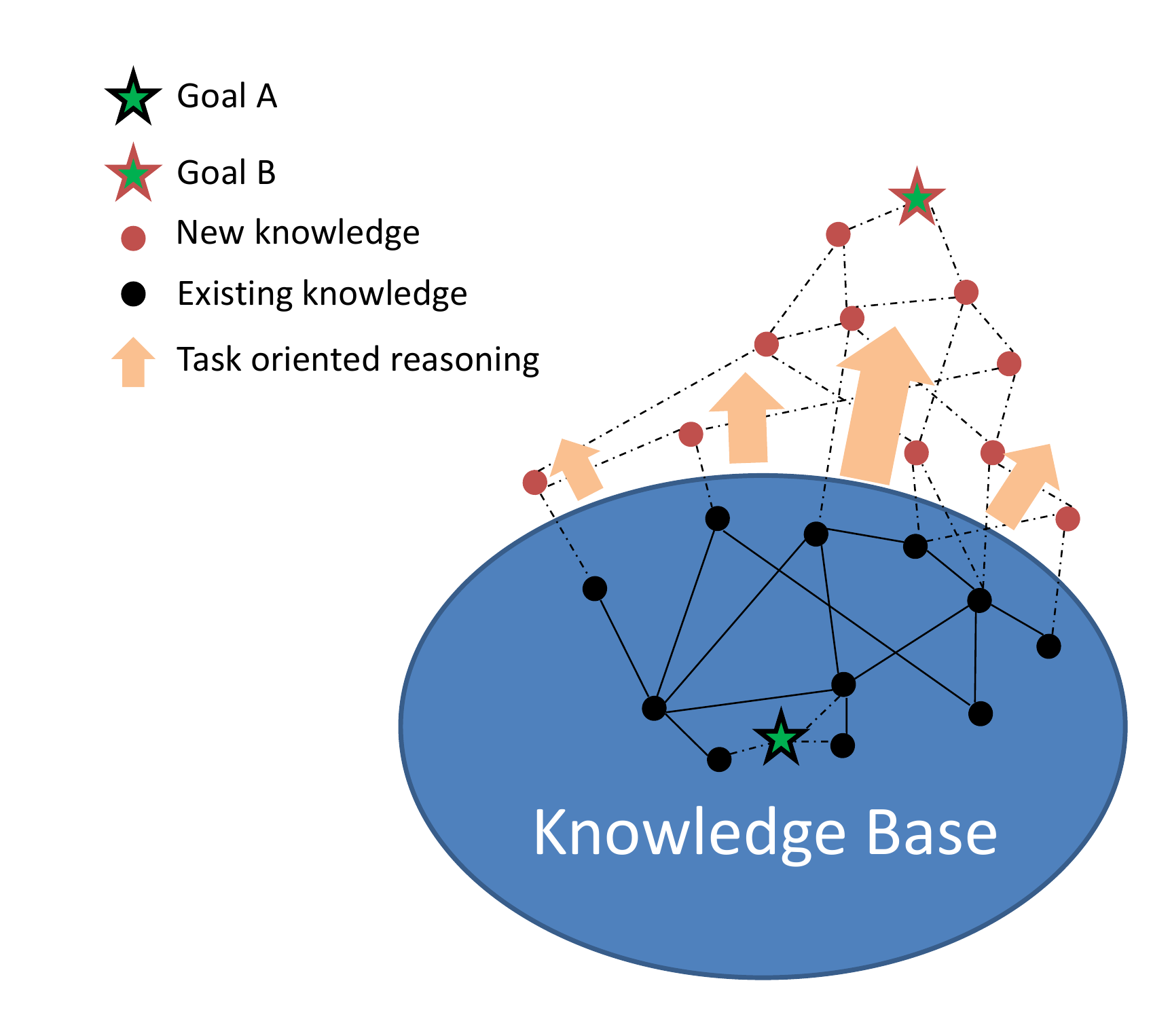}}
		\caption{Reaching Goal A inside the existing Knowledge Base can be achieved without any reasoning by dense sampling of the space. To reach Goal B, the system has to apply knowledge processing mechanisms creating new knowledge and skills from existing ones.}
		\label{Tasks oriented reasoning
		}
	\end{center}
	\vskip -0.2in
\end{figure}

Indeed, a fixed knowledge system will have many limitations to perform in rich scenarios like the real world. For example, it will need to know beforehand all the exceptions to a rule, since those cases cannot be judged by their similarity with previous data. It can only solve problems in the limited space covered by the existing knowledge which contradict the commonly accepted definition.

On the other hand, intelligent systems can increase their knowledge towards the area containing the task's solution. The goal of AI is finding architectures capable of doing goal-oriented learning (Figure 1).

Under this vision, testing for intelligence requires information about the limits of the existing knowledge in the system; which can be the case if the evaluator has access or supplies the training data. Examples of testing the capacity to acquire new information could be changing the grid size of board game, colors, shapes and other properties to values never seen before. The degree of intelligence is measured by the capacity to construct new knowledge, using the data manipulation tools of the system, reaching targets further away from the initial knowledge (e.g. a new image can be recognized as a rotated and scaled version of a known one).

Side by side performance comparison (e.g. using SOTA metrics) in a single task should not be used as a comparative measure of intelligence between two systems. There is an infinite amount of intellectual tasks that humans cannot accomplish in isolation, due to constraints in terms of memory, computational precision and speed. We rely on tools that assist us in solving such problems. Overall, general systems will be outperformed by systems specialized in a task. This is not an indication of higher intelligence since the approach of general systems is to create and rely on specialized ones to accomplish tasks. Equally, the development of AI systems should not focus on mastering specialized tasks but in developing their capacity to combine the existing knowledge optimally towards any given goal; decomposing the goal into subtasks and flexibly coordinating the solution of those subtasks.

Systems with fixed knowledge and specialized systems in general, can have significant practical value (e.g. in medical imaging diagnostics) performing tasks with high speed and accuracy. Deep neural networks (DNNs) with fixed knowledge do intelligent data processing at training time and, after deployment, imitate the behavior of similar training inputs. Here we focus on the limitations of such systems for generalization in broad scenarios like the real world. 

We have concentrated on the properties of AI as problem solver but, thanks to the recent attention to ethical AI and the AI alignment problem \cite{Leike18}, there is a growing awareness about the need to consider, from early stages, the properties of AI systems as  tools. The desired properties in each case are listed below.

Desired properties of the system as a problem solver: \emph{common sense (acting according to contextualized experience), unsupervised learning, zero-shot learning, continuous learning, knowledge transfer, task partitioning, creativity and learning from abstract knowledge}.

Desired properties of the system as a tool: \emph{interpretability, transparency, traceability, stability \& robustness, reproducibility, self-monitoring, maintainability, controllability, know when doesn't know, value based behavior, knowledge extraction, bias detection and fairness}.
 
Our vision of the AI development methodology implies regularly checking the product  versus the listed properties; ensuring its compliance by design or by specially crafted tests.
\section{Learning Grounded Representations}
Let’s imagine a typical learning toy for babies: a board with shapes to be filled with the right piece. If we show to the system a square shaped piece without showing the board: what information should be retained about the object? Our expectation is that, when confronted later with the task of assembling the pieces, the system will associate the right shape on the board with the previously seen object.
\begin{figure}[ht]
	\vskip 0.4in
	\begin{center}
		\centerline{\includegraphics[width=\columnwidth]{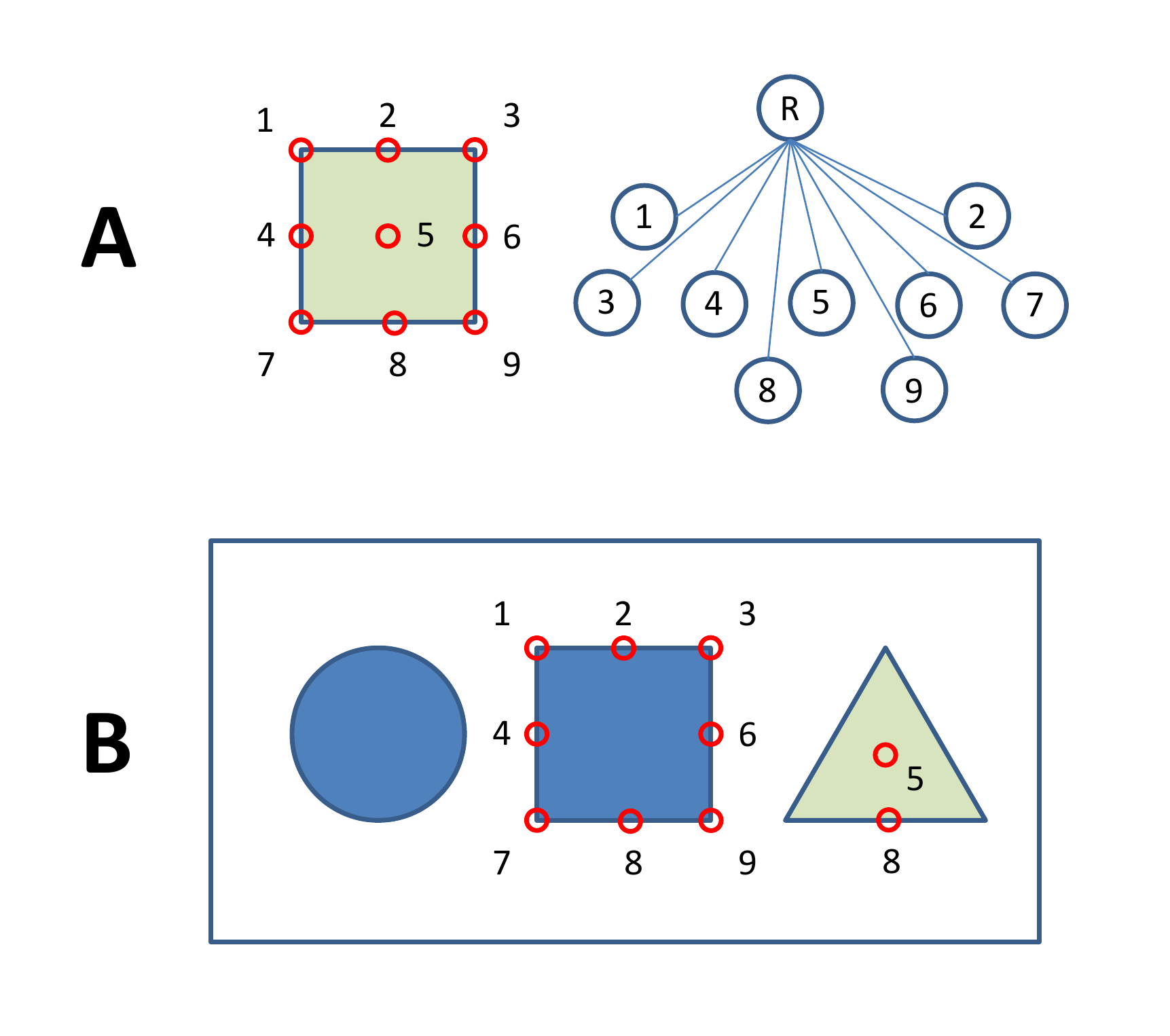}}
		\caption{Learning should extract all relevant features of an input. \textbf{A}. The representation \textbf{R} of the square contains all detected features (\textbf{1}-\textbf{9}). \textbf{B}. Similar features detected in the board activate the associated shape. The square in the board contains all detected features except for the color (\textbf{5}), which is associated from the triangle as well as the bottom horizontal line (\textbf{8}).}
		\label{Tasks oriented reasoning
		}
	\end{center}
	\vskip -0.2in
\end{figure}

Since we may need \emph{any} of the properties of an input to be used \emph{later} while solving a task; we will need to extract \emph{all} properties at \emph{learning time}. That would imply, in the example, creating an internal representation of the shape, color, texture and other properties of the piece (Figure 2). Moreover, it should be possible to locate the high level concept that contains any specific property. Consequently, the created representation cannot be an entangled combination of all properties. 

This leads to the principle:

\textbf{Principle I. Mirrored Representations}: \emph{The internal representation stores information in a hierarchical compositional structure that matches the input in a generative way}.

Intuitively, it is easy to understand that every property or feature we can identify in one input (e.g. an object) becomes attached to its representation (e.g. object’s size, shape, etc.). The hierarchical compositional representation implies that given the object we can activate the feature and given the feature we can activate the object. We rely on this link when we use an object to accomplish a task. Beyond shared properties, we need also the same coherence; and finally, we need a mechanism to build such representation.

Creating a representation is, in practice, dependent of input's inductive priors (e.g. simple Hebbian association for spatial inference with similar activation patterns associated under the same parent node). Every time a parent node is created it propagates a signal downstream that is contrasted with current activation. This cancellation removes the child nodes from the pattern seeking candidates. Synchronized loops of bottom-up activation followed by top-down cancellation create a compositional representation that is generative and that recreates the input top-down.

The sketched learning mechanism meets several of our desired properties: it is unsupervised, learns from a single input and allows continuous learning, reusing without distortion previous knowledge. Also, the coherence of the input is embedded in the representations and its features are represented at scale: bigger features are detected first and have more influence than subtle ones. This makes the system stable to small perturbations like the ones used in adversarial attacks.

Incremental learning and knowledge reuse are other important properties for reasoning; supporting transitivity, knowledge transfer and partial information completion.

This approach takes a different strategy to learning compared to deep neural networks. Here we start from an empty network while DNNs are created as multiple layers of connected nodes passing the activation from input to the output layer. In such formulation, DNNs already contain unfounded `innate knowledge' at inception time since they are able to produce an output for any input. This knowledge is adjusted during training to reflect the regularities of the data but residual behavior may remain leading to inappropriate responses to non encountered patterns. 

In this proposal, by design, unknown patterns cannot trigger existing high level nodes; while in the case of DNNs it is possible, showing that they don't know when they don't know.  
\section{Problem Solving and Logical Operations}
Autonomous task execution is a main objective in creating intelligent systems. The task can vary from simple static input recognition till multifactor decision making in dynamic environments.

DNNs specify the goal using a loss function and RL uses rewards provided by the environment to favor specific behaviors. Both methods require heavy human engineering and are not generalizable to all tasks.

In this framework we will supply the goal to the system using its regular low level representation. People use the same approach when solving a task by creating a mental picture of the target state. This way, to assemble a puzzle we supply to the system the final picture; if we want to exit a maze, we show the controlled agent at the exit. It is up to the system to convert the input into actionable information. 
\begin{figure}[ht]
	\vskip 0.2in
	\begin{center}
		\centerline{\includegraphics[width=\columnwidth]{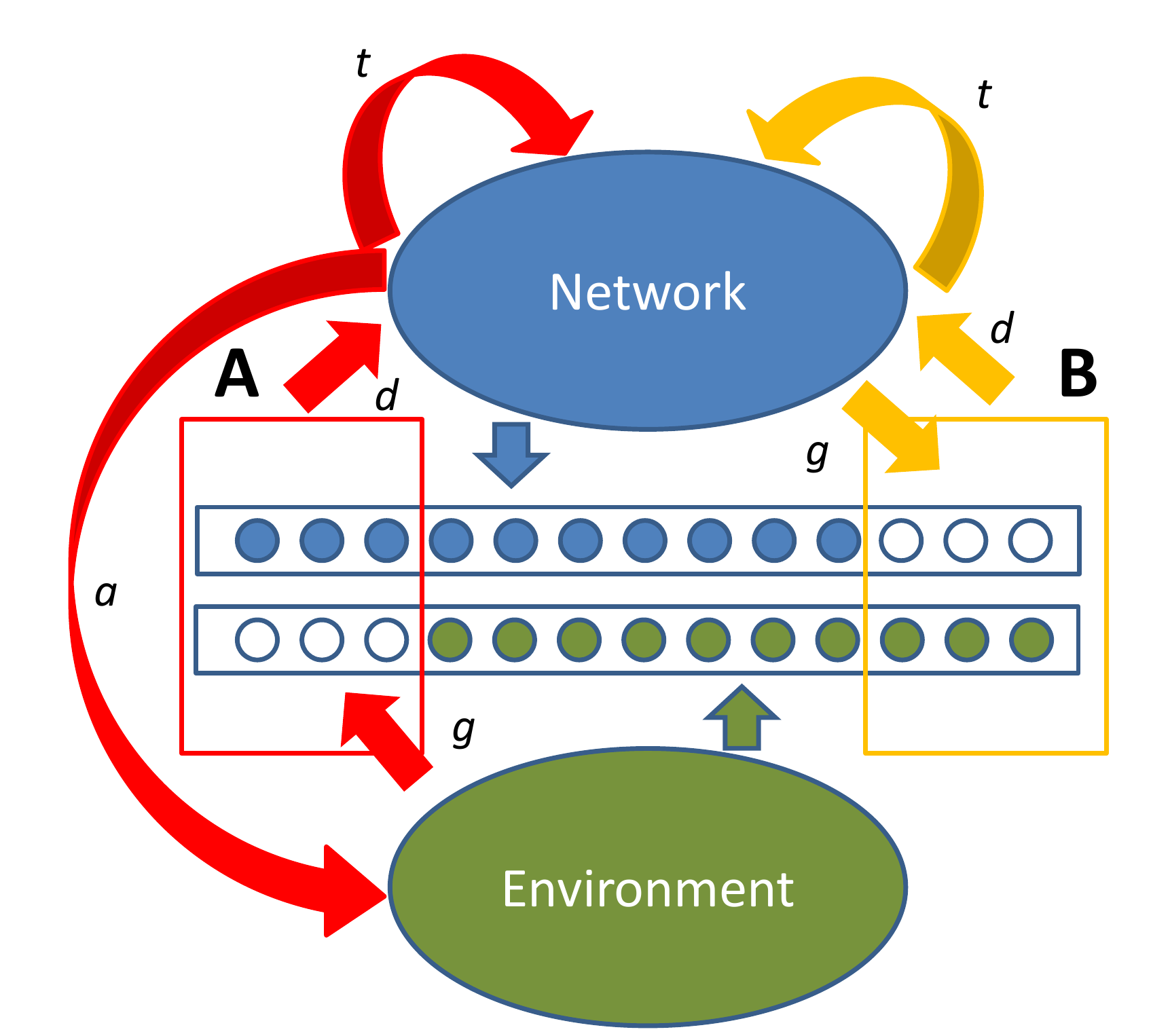}}
		\caption{Grounded Actions: the resulting activation of the bottom layer -computed as input (green) versus internal representation (blue)- is the mechanism of information processing. \textbf{A}. \emph{Problem solving} (red) takes place when the internal representation is not matched by the input. \textbf{B}. \emph{Learning} (yellow) happens when the input has no matching representation. The discrepancy \textbf{\emph{d}} between the input and the representation is propagated into the system, triggering transformations \textbf{\emph{t}} and actions \textbf{\emph{a}} to achieve the goal \textbf{\emph{g}} of filling the blanks.}
		\label{
		}
	\end{center}
	\vskip -0.2in
\end{figure}

We will extend the mechanism used for learning to the problem solving case, when the internally activated representations are not matched by the external inputs. In this situation the system is motivated to find the equilibrium by other means: by changing the world so that the external inputs match the internal representation (Figure 3). 

Actions and transformations are functions that modify the output representation. By performing concrete actions in the environment (e.g. pushing an object or rotating an image) the system associates such action with the resulting change in the input; transformations are generalizations (abstractions) of the perceived change from actions. During problem solving, the system triggers actions after finding the right set of internal transformations for the goal.

The unified principle of information processing is then formulated as:

\textbf{Principle II. Grounded Computations}: \emph{Bottom level activations drive information processing in the system. Restoring the equilibrium at the bottom layer leads to learning, problem solving and imagination}. 

Learning takes place when the input cannot be expressed using existing patterns and new patterns need to be created for matching such input. Problem solving consists in finding the actions in the environment that would cause a change in the input so it matches the given internal representation.

While we usually see a clear distinction between learning and problem solving in ML literature, in practice, the two cannot be completely separated. While learning, we frequently have to solve the problem of expressing the given input as transformations of the existing knowledge (e.g. a known object is presented rotated). Equally, during problem solving we have to manipulate new combinations of concepts (e.g. a repeated sequence of actions) that are learned and represented in a new concept for further processing. 

We need to stress the importance of the iterative nature of information processing described in this principle. The bottom-up and top-down loops will continue while there is activation mismatch at the bottom layer. As explained later, loops in information processing are required for reasoning. Systems with only one possible path per world state, like some DNN solutions, need to store a large number of possibilities and cannot benefit from generalization or ad hoc reasoning. For example, in the scenario of wearing special image inverting lenses, it is enough for a system to find the inverse image transformation to continue reusing all existing knowledge. Single state-to-action models need to be trained in all possible input transformations, which becomes intractable.
\subsection{Inhibition and Reasoning}
Reasoning is a cornerstone of intelligence. It reflects the iterative nature of thinking where a logical sequence of arguments leads from one assertion to another or cause a restart of the analysis with different set of assumptions. 

Implementing explicit and traceable reasoning steps in neural networks is a significant challenge. Evaluations of abstract reasoning using DNNs \cite{Barrett18} lack the essential transparency of the process and achieve very limited results. 

Here we construct reasoning as a network wide logic process performed at three levels: 
\begin{labeling}{\textbf{Logic Interaction}}
	\item[\textbf{Logic Interaction}] Bottom level. Conditioning the behavior of one element given the state of another.
	\item[\textbf{Common Sense}] Middle level. Representing the system wide impact of all interactions.
	\item[\textbf{Counterfactuals}] Top level. Changing the preconditions of a group of elements in the system and restarting the deductive process.
\end{labeling}

\subsubsection{Logic Interaction}
The presence of one element may increase or decrease the likelihood of the presence of another; in the extreme case, the presence of one element implies the presence or absence of the second. In the logical domain, we can say that an element is present or `true' in a given context when it triggers in that context as result of its activations. Equally, the inhibition of an element indicates its negation, i.e. the element is `false' if prevented from triggering in the context.

Imagine using the Hebbian rule to enhance the positive association between two elements that fire together. In order to do logic operations we need to reflect also the interaction between contradictory statements. The following extension of the Hebbian rule will be used: \emph{if the occurrence of one element takes place at the same time as the removal of another, then an inhibitory link is established between them}. Indeed, in Hebbian terms, the first event is positively linked to the negation of the other, i.e. its inhibition. 

Changes in the firing capabilities of elements influence information processing in the system, which is reflected in the following principle:

\textbf{Principle III. Adaptive Logic}: \emph{The computational logic of the system is constantly changing according to the state of elements in the network.}

In general, many ad hoc factors can modify the firing capabilities in the system; in the case of inhibition, the following particular rules are formulated:

\emph{\textbf{Principle III. Adaptive Logic (Inhibition)}: A) Parent nodes of an inhibited node are also inhibited. B) If all higher level abstractions containing a given element are inhibited then the element is also inhibited. C) In the state/action space, if a given non-target state can only do a transition to inhibited states, then such state is also inhibited}.

All necessary logical operations for reasoning can be obtained by combining activation and inhibition, adding the benefits of classical logic-based processing to the neural networks context. E.g. if two mutually exclusive concepts, like `water' and `fire', are in the same location in a context, the interpretation of the input will depend on the logical interactions between them (Figure 4). 
\begin{figure}[ht]
	\vskip 0.2in
	\begin{center}
		\centerline{\includegraphics[width=\columnwidth]{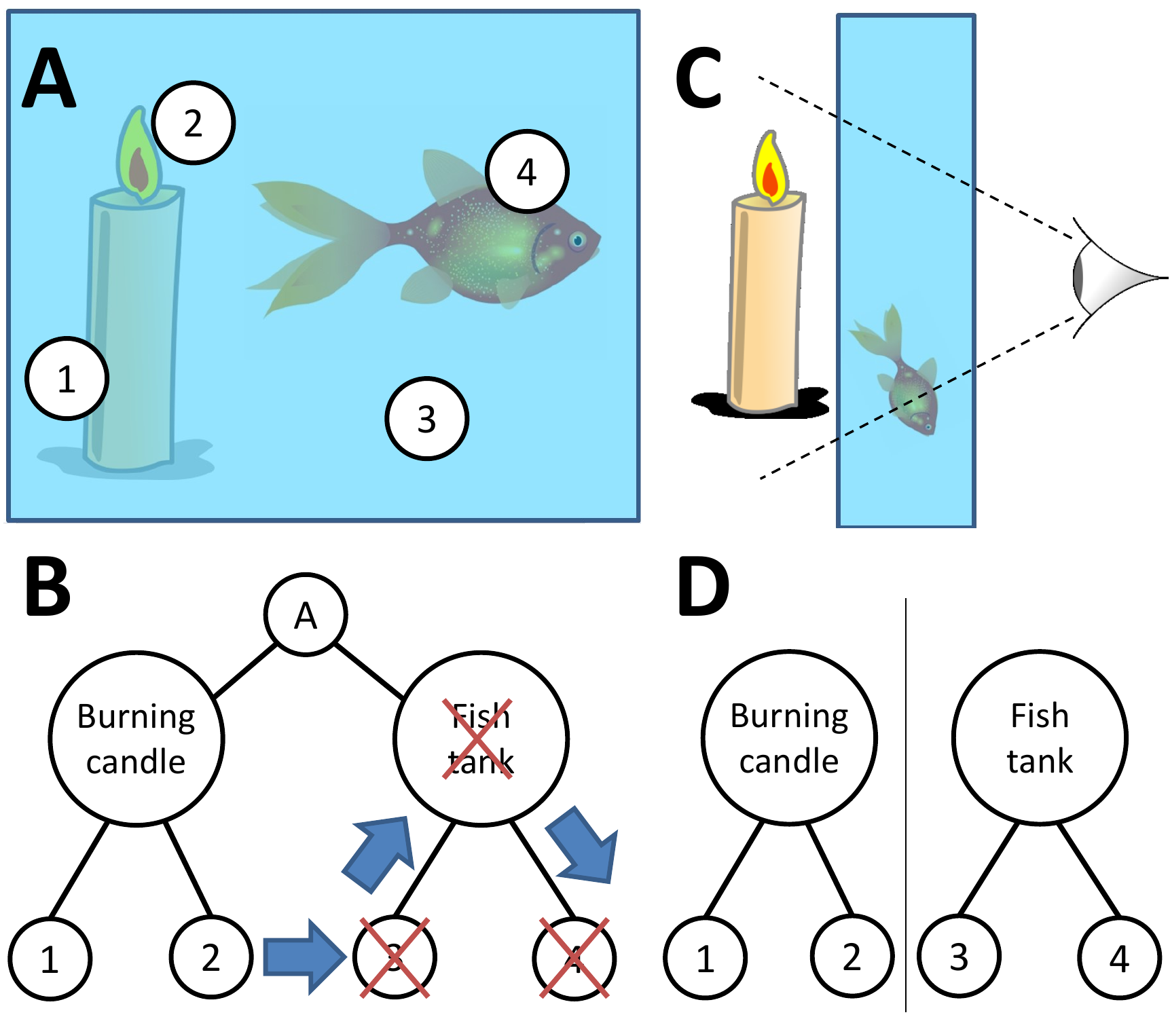}}
		\caption{Logical Interaction. \textbf{A}. A burning candle (\textbf{1}, \textbf{2}) seems to be in a fish tank (\textbf{3}, \textbf{4}). \textbf{B}. If the system has learned that \textbf{2} and \textbf{3} are mutually exclusive, so if \textbf{2} triggers it implies that \textbf{3} is inhibited, leading to the inhibition of the parent and therefore of \textbf{4}. Consequently, an alternative explanation to a fish tank is required. Similar reasoning can be done assuming \textbf{3} to be true. \textbf{C}-\textbf{D}. The search for explanations will continue until all low level representations (\textbf{1}-\textbf{4}) are explained in the example.}
		\label{
		}
	\end{center}
	\vskip -0.2in
\end{figure}
\subsubsection{Common Sense}
Maze and exploratory challenges are frequently used to evaluate the decision making capabilities of AI systems. The preferred algorithm for such scenarios is RL, based on rewards to obtain state-to-action policies. RL agents require a large amount of iterations to improve their behavior on the environment due to the lack of common sense, which leads to the execution of spurious actions like, e.g., moving several times over the same sections of the path. 

In fact, maze-like scenarios can be solved in nature even by unicellular organisms \cite{Nakagaki01}. We know also that animals create an internal representation of the explored path to be used later for decision making \cite{Bush15}. 

In this framework, by applying the already described principles we achieve, in an unsupervised manner, a performance equivalent to the one expected from a rational agent. According to the learning method, at every step, a higher level concept is created including the previous state (canceled) plus the new position; in crossroads, a new high level element is created for each direction. When reaching a dead end (a place where all directions are inhibited) the state is inhibited (Figure 5). 
\begin{figure}[ht]
	\vskip 0.2in
	\begin{center}
		\centerline{\includegraphics[width=\columnwidth]{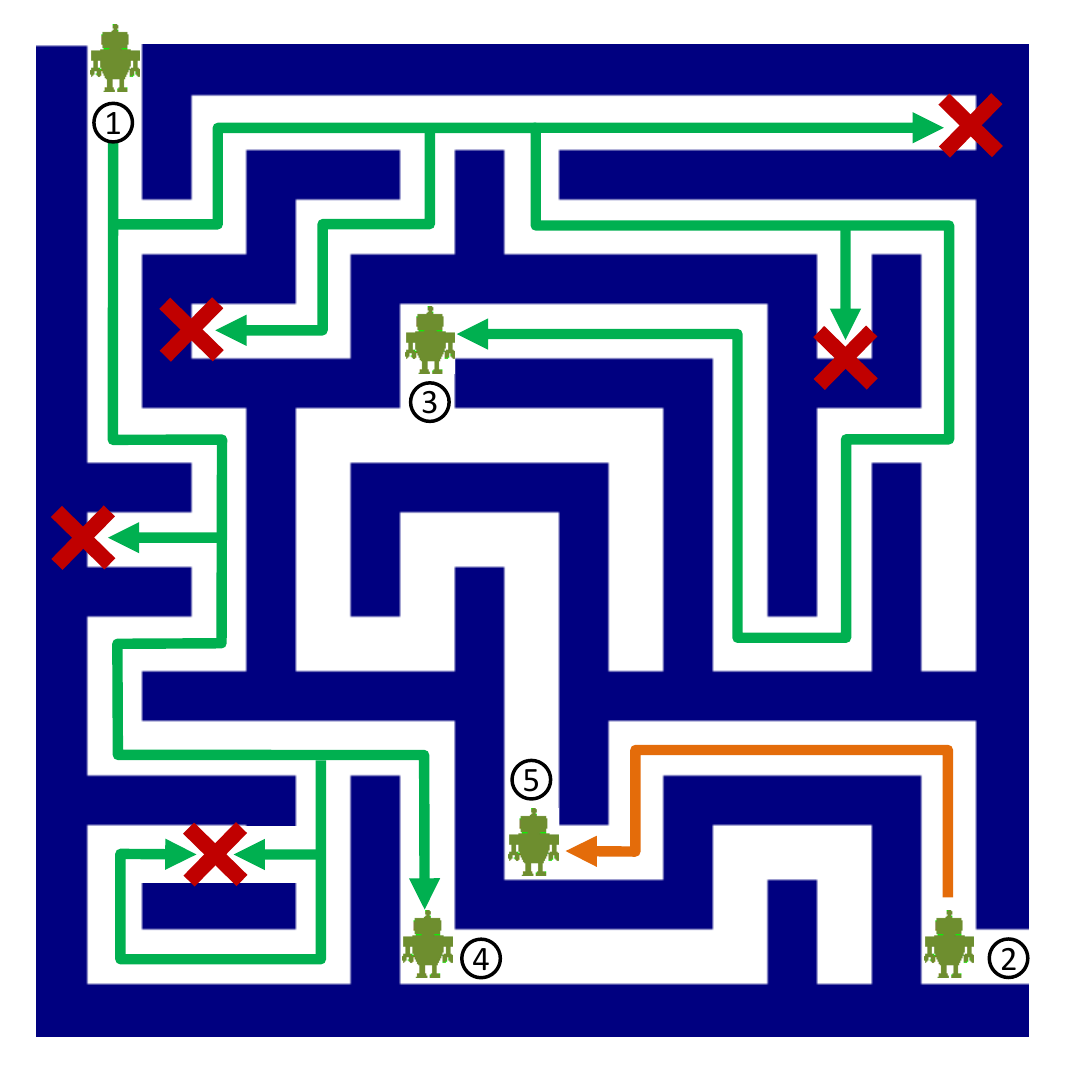}}
		\caption{Common Sense. The agent starts from location \textbf{1} and has to reach exit \textbf{2}. Directions reaching dead ends or already transited paths are inhibited (red). Currently active paths are \textbf{3}, \textbf{4} and current target is \textbf{5} by abductive reasoning.}
		\label{
		}
	\end{center}
	\vskip -0.2in
\end{figure}

Contrary to DNNs, in this approach the system remembers the history of each path and uses this information for decision making. At the same crossroad, different directions may be taken depending on whether any trajectory passed by that location already. A high efficiency is achieved thanks to parallel computation and sharing information between alternatives; which is the type of behavior we observe in the brain. Another efficiency improvement is obtained by the exploration backwards from the goal. Humans apply this method regularly during problem solving but is not applicable to solutions based on function optimization.
\begin{figure*}[ht]
	\vskip 0.2in
	\begin{center}
		\centerline{\includegraphics[width=\textwidth]{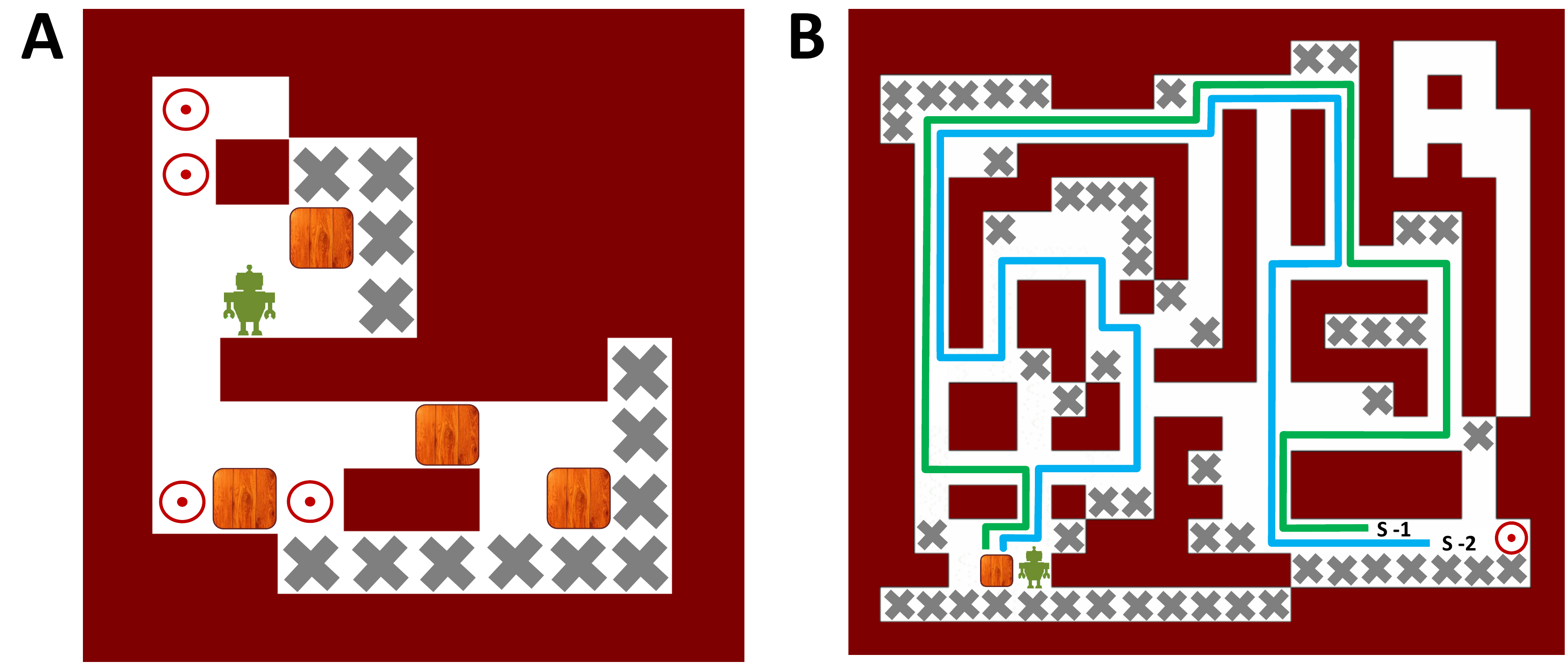}}
		\caption{Common sense reasoning removes deadlock states making then inaccessible to the planning (gray crosses). Even in scenarios considered difficult by RL implementations (\textbf{A}, \textbf{B}) the resulting space of actions can be covered with few logical paths. To find an alternative solution \textbf{S-2} to the existing \textbf{S-1} we restart the process after inhibiting \textbf{S-1} and, consequently, all paths attached only to it.}
		\label{
		}
	\end{center}
	\vskip -0.2in
\end{figure*}

This principle allows not only finding the solution to the problem but, equally  important, detecting the cases where there is no solution. If all paths are inhibited without reaching the goal, that means that there is no solution. RL evaluations are focused on scenarios where the solution exists and the agent is guided towards the goal using rewards. In reality, we only know the goal but ignore how to reach it or whether it is reachable at all in the environment; such general case is solved in this framework.
\subsubsection{Counterfactuals and the Power of Inhibition}
As mentioned earlier, common sense reasoning with neural networks can quickly find a solution to a maze or conclude that there is no solution possible. More advanced scenarios, like Sokoban, represent a challenge for current state-of-the-art ML solutions \cite{Guez19,Kansky17} where their performance is expressed in partial success rates. The reasoning principles described so far reduce the complexity of such scenarios removing the logically inaccessible areas (Figure 6) and guarantee finding the solution. By using counterfactuals we can also cover the full space of possible solutions. 

An inhibited high level element propagates downstream the inhibition to the lower level elements exclusively linked to it, resulting in such high level element `disappearing' from the network. This creates the possibility of using inhibition for: finding alternative solutions, controlling system's behavior and improving efficiency. Indeed, to find an alternative solution it is sufficient to inhibit found ones at their high level representation and restart the task. The process can be repeated until no more solutions are found. Similarly, if we have a large task where multiple subtasks have known solutions, by inhibiting those subtasks we direct computational resources to the unresolved problems; as done regularly by people. Finally, inhibition enables effective control over the system by allowing only necessary access to resources; e.g. an agent with low privileges will face only allowed actions. The restrictions extend from the resources of interest to the action space; e.g. if we inhibit the concept of a broken vase to prevent the event from happening them a vase located in a corridor may change the action space of the cleaning robot so that entering that corridor is not possible. 

This dynamic network-wide reconfiguration differs from the traditional agent-centric control. It can only be implemented in systems with hierarchical representations allowing direct access and configuration of high level concepts. Inhibition seems to play also a relevant role in human psychology where actions and thinking processes appear to be conditioned by restrictions imprinted in the world model.
\section{The AI Model and Neuroscience}
Applying the mechanisms of biological intelligence in AI is a reasonable strategy considering that the brain is the best reference of a flexible and general problem solver. Also, we expect that, by creating such models, we can gain insights about the functioning of our thinking processes. 

Besides the already mentioned similarities with brain's functioning, the model has other common points with neuroscience evidence:
\begin{itemize}
\item Bottom-up and top-down loops can be related to brain waves keeping high and low level representations in sync.
\item Intuition matches the bottom-up consensus from many sources. There is evidence that having short decision times (i.e. not sufficient to engage in reasoning loops) people decisions are similar to the ones from feedforward models \cite{Elsayed18}.
\item Awareness as top-down integration containing only the attended information and reflecting the taken bottom-up decisions. The delayed temporal correlation agrees with neuroscience evidence \cite{Dehaene11}.
\item Matches the massive parallelism and synchronized serialization.
\item Matches the fact that internal representations are generative as we know from dreams.
\item Correlates with perceived experience of imagination as a chained stream of associations.
\item It is aligned with the evidence that inhibition and excitation signals are present in the same order of magnitude.
\item Explains and integrates the roles of curiosity, task partitioning and internal goals into a single mechanism.
\end{itemize}
\section{Discussion}
A model of intelligence is presented based on few principles:
\begin{enumerate}[I]
	\item \textbf{Mirrored Representations}. The \emph{what}. Compositional representations are used for capturing input's coherence and implementing, thanks to components' reuse, knowledge transfer and partial knowledge completion.
	\item \textbf{Grounded Computations}. The \emph{how}. This is the core motivational engine driving computations by top-down explanation of low level activity. This principle implements task partitioning, curiosity and imagination.
	\item \textbf{Adaptive Logic}. The \emph{why}. It implements the core logic needed for information consistency, processing priorities and alternative interpretations. With common sense we narrow the space of candidate world states (i.e. take better decisions) and with counterfactual reasoning we expand this space.
\end{enumerate}
These principles aim at building structured representations that capture the spatial, temporal and logical interdependence between the abstractions found in the input. Consequently, the activation of few elements may lead to iterative, equilibrium seeking, dynamics propagated over the network.

We claim that the proposed solution meets in its design key desired properties as problem solver (\emph{e.g. common sense, unsupervised learning, creativity}) as well as a tool (\emph{e.g. interpretability, transparency, controllability}). Also, thanks to the use of grounded representations, which keep independent information stored separately, it is possible to use multiple data types and learning different tasks in a single network without interference. Grounded representations are associated here with the concept of understanding, becoming particularly important in areas like NLP where symbols already include abstracted information. All mentioned expectations of the model will be measured in the ongoing implementation using specific tests. 

There are important high level differences between this paradigm and main ML solutions. The goal, reflected by the core computing algorithms, is in DL and RL to solve the given task, while in our proposal is to create a model of the environment. 

In DL and RL the system is driven by task specific motivations included in the loss function or in the rewards. Additional techniques are used to stabilize the behavior in face of context's variability. In that sense, generalization is impacted by the task specific focus, the handcrafted functions and the tuning to the environment. This correlates with the fact that excellent performance is seen only in single tasks, with human in the loop and using simplified virtual environments.
 
In this proposal we focus on the general task of achieving equilibrium between the internal model and an arbitrary environment; using unsupervised learning or actions to mitigate the discrepancies between the two. We solve a task by inserting its goals in the internal model as representations of desired states of the environment.

We believe that iterative reasoning is required for extracting the maximum value from small data and dealing with unexpected deviations. In the real world critical information may be collected from rare events or from experiments too expensive to be repeated many times. The reasoning and continuously adapting capabilities of this solution makes it a good candidate for autonomous robot exploration, drug discovery and other applications.
\section{Conclusions}
Intelligent machines will be crucial for solving complex problems in science, technology and society. We consider that the scarce progress towards general AI is mainly due to the lack of good theoretical models and not of data or computational resources. We aim at building AI systems that meet expectations as problem solvers and as tools, while capturing the complexity and power of biological intelligence. The presented theory of intelligence attempts to make progress in a systemic approach based on practical, explained and provable statements. We are certain that a continued effort on refining and extending these core principles together with a matching implementation will lead to more comprehensive and powerful intelligent systems. 
% In the unusual situation where you want a paper to appear in the
% references without citing it in the main text, use \nocite
\nocite{langley00}

\bibliography{Computational_Principles_GI}
\bibliographystyle{icml2019}

\end{document}